\documentclass[journal]{IEEEtran}
\usepackage[pdftex]{graphicx}
\usepackage{amsmath,epsfig,multirow}

\ifCLASSINFOpdf
\else
\fi
\begin{document}
\title{Technical Report for Valence-Arousal Estimation in ABAW2 Challenge}

\author{Hong-Xia~Xie, I-Hsuan~Li, Ling~Lo, Hong-Han~Shuai, and~Wen-Huang~Cheng %,~\IEEEmembership{Member,~IEEE,}
        %John~Doe,~\IEEEmembership{Fellow,~OSA,}
        %and~Jane~Doe,~\IEEEmembership{Life~Fellow,~IEEE}% <-this % stops a space
% \thanks{M. Shell was with the Department
% of Electrical and Computer Engineering, Georgia Institute of Technology, Atlanta,
% GA, 30332 USA e-mail: (see http://www.michaelshell.org/contact.html).}% <-this % stops a space
% \thanks{J. Doe and J. Doe are with Anonymous University.}% <-this % stops a space
% \thanks{Manuscript received April 19, 2005; revised August 26, 2015.}
}

\maketitle

\begin{abstract}
In this work, we describe our method for tackling the valence-arousal estimation challenge from ABAW2 ICCV-2021 Competition. The competition organizers provide an in-the-wild Aff-Wild2 dataset for participants to analyze affective behavior in real-life settings. 
We use a two stream model to learn emotion features from appearance and action respectively. 
To solve data imbalanced problem, we apply label distribution smoothing (LDS) to re-weight labels.
Our proposed method achieves Concordance Correlation Coefficient (CCC) of 0.591 and 0.617 for valence and arousal on the validation set of Aff-wild2 dataset.
\end{abstract}

%\begin{IEEEkeywords}
%IEEE, IEEEtran, journal, \LaTeX, paper, template.
%\end{IEEEkeywords}

% \fi
%\IEEEpeerreviewmaketitle

\section{Introduction}
% 1. 先介紹emotion recognition的應用, valence, arousal
\IEEEPARstart{P}{eople} study facial expression recognition research for a long history, but still have some challenges to be addressed, especially in-the-wild scenarios. 
Current in-the-wild dataset contains limited human emotion annotation due to the cost and time requirement. And this causes the limitation of multi-task method progression and applications for emotion recognition in real life. Recently, to tackle such problems, Kollias et al. \cite{kollias2020analysing}, \cite{kollias2019expression}, \cite{kollias2019face}, \cite{zafeiriou2017aff}, \cite{kollias2021affect}, \cite{kollias2021distribution}, \cite{2106.15318} held Affective Behavior Analysis in-the-wild (ABAW2) ICCV-2021 Competition and built the large-scale Aff-Wild2 dataset, which includes annotations of valence/arousal value, action unit (AU), and facial expression for three different recognition tasks.
% \cite{kollias2019expression}, \cite{kollias2019face}, \cite{kollias2019deep}, \cite{zafeiriou2017aff}, \cite{kollias2017recognition}
Valence represents how positive the person is while arousal describes how active the person is. AUs are the basic actions of individuals or groups of muscles for portraying emotions. As for facial expression, it classifies into seven categories, neutral, anger, disgust, fear, happiness, sadness, and surprise.

\begin{figure}
  \centering
  \includegraphics[width=0.5\textwidth]{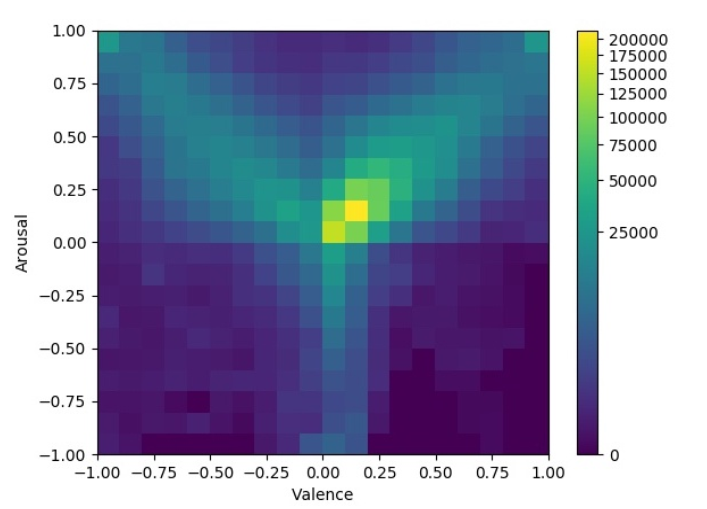}
  \caption{The 2D Valence-Arousal Histogram of Aff-Wild2~\cite{kollias2021analysing}
  \label{fig:va_histogram}}
\end{figure}

% 2. 介紹valence arousal資料存在imbalance問題
However, the annotation of valence and arousal in Aff-Wild2 dataset is distributed imbalance heavily, as shown in Fig.~\ref{fig:va_histogram}. Deng et al.~\cite{deng2020multitask_FG1st} addresses the data imbalance problem by importing external AFEW-VA dataset~\cite{kossaifi2017afew} which contains 30,051 frames followed by rebalancing. However, the distribution of the resampled dataset is still not fully balanced.
% ~\cite{deng2020multitask_FG1st}
% We downsample the Aff-wild2 dataset by 5, and merge it with the AFEW-VA dataset. We then discretize the valence-arousal scores into 20 bins of the same width. We treat each bin as a category, and apply the oversampling/undersampling strategy as used in the expression recognition task. The distributions of the valence-arousal scores in the downsampled Aff-wild2 dataset, the AFEWVA dataset and the merged dataset are shown in Figure 2 (c). The merged dataset is resampled to improve balance, but because of some rare cases, e.g., (V,A) = (1,−1), the distribution of the resampled dataset is not fully balanced.

% However, training on more datasets also introduces large training efforts.
% from DIR paper
All imbalanced learning methods, directly or indirectly, operate
by compensating for the imbalance in the empirical label
density distribution. This works well for class imbalance,
but for continuous labels the empirical density does not accurately reflect the imbalance as seen by the neural network.
Hence, compensating for data imbalance based on empirical
label density is inaccurate for the continuous label space~\cite{DIR}.
% 3.我們在此提出一個什麼樣的解決方案
In this work, we propose a two stream valence-arousal estimation network based on MIMAMO Net \cite{deng2020mimamo}.
% The model, shown in Fig. 1, gets state-of-the-art performance on the OMG \cite{barros2018omg} and Aff-Wild \cite{kollias2019deep} dataset.
The spatial and temporal learning are used to capture appearance and facial action information, which can further improve emotion recognition. Moreover, to prevent data exhibit skewed distributions, we use LDS, which were proposed by Yang et al. \cite{DIR}, for dealing with continuous targets to preserve a uniform distribution.

\section{RELATED WORK}
In recent years, most of the existing research for facial expression recognition focused on valence-arousal estimation, facial action unit detection, and expression classification. We will introduce the latest related work of valence-arousal estimation study.

Many data are in laboratory settings. However, models that perform well on controlled conditions don't necessarily work well on uncontrolled ones. The ideal is still far from reality. Therefore, in-the-wild datasets come to exist. Kossaif et al. \cite{kossaifi2017afew} proposed a new dataset called AFEW-VA and found that geometric features performed well no matter what settings were. But it was unuseful for dynamic architecture since some of the clips in the dataset were too short to explore information between frame and frame. Barros et al. \cite{barros2018omg} proposed OMG dataset, collected from YouTube in real-world settings. The main keyword to select the videos was "monologue." Kollias et al. \cite{kollias2019deep} built a large-scale Aff-Wild dataset, collected from Youtube, and proposed deep convolutional and recurrent neural architecture, AﬀWildNet. A CNN extracted features while an RNN aimed to capture temporal information. Furthermore, their works not only got high performance on dimensional aspects but also for expression classification. 

Chang et al. \cite{chang2017fatauva} proposed an integrated deep learning framework that used the concept of applying the information of facial action unit detection to estimate valence-arousal intensity. They had shown that exploring the relationship between AUs and V-A was helpful for V-A research. Pan et al. \cite{pan2019deep} proposed a two-stream network to utilize effective facial features. The model contained CNN and LSTM. For temporal stream, the former extracted temporal features; the latter resolved the temporal relation between frames. For spatial stream similar to temporal one, the former extracted spatial features; the latter analyzed the spatial association between frames. Kim et al. \cite{kim2021contrastive} tackled regression problems with adversarial learning, which enabled the model to better understand complex emotion and achieved person-independent facial expression recognition. Also, they proposed a contrastive loss function and improved the performance effectively. This study proved the potential of adversarial learning instead of conventional methods on emotion recognition.

% \begin{figure*}
%   \centering
%   \includegraphics[width=1\textwidth]{model.jpg}
%   \caption{MIMAMO Net \cite{deng2020mimamo}}
% \end{figure*}

\section{METHODOLOGY}
\subsection{Overall Architecture}
\begin{figure}
  \centering
  \includegraphics[width=0.48\textwidth]{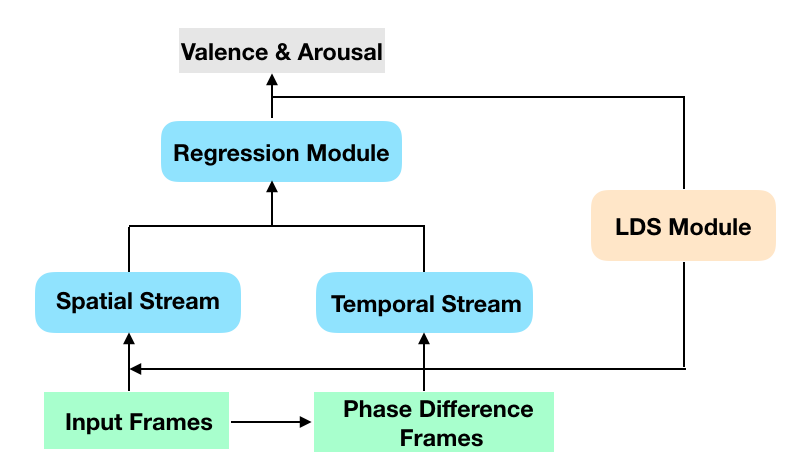}
  \caption{The architecture of our method.}
  \label{fig:architecture}
\end{figure}
\begin{figure}
  \centering
  \includegraphics[width=0.5\textwidth]{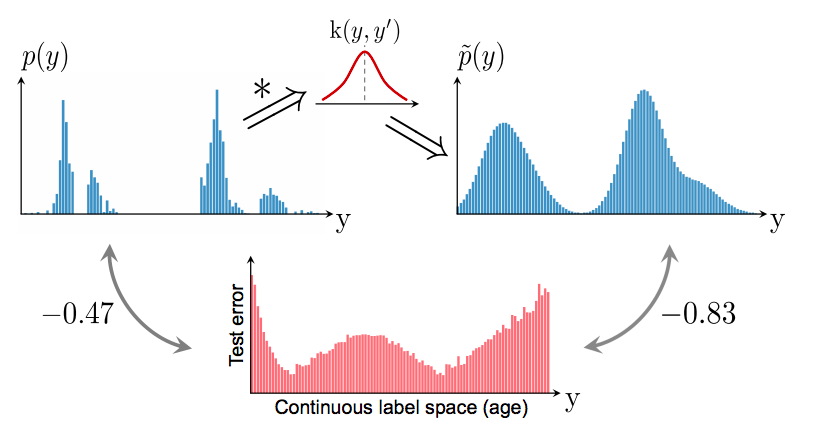}
  \caption{Label distribution smoothing (LDS)~\cite{DIR}.
 }
  \label{fig:LDS}
\end{figure}

Inspired by~\cite{deng2020mimamo}, we build a two-stream network for valence and arousal estimation, as shown in Fig.\ref{fig:architecture}.
% Phase difference images
Specifically, for input images, we first compute their corresponding phase difference images. 
% spatial stream
The spatial stream uses the pre-trained ResNet50 model to extract features of the pool5 layer, then the feature vector are fed into a MLP module to get the final feature vector. 
% temporal stream
While in the temporal stream, it utilizes phase difference images to obtain the relationship among frames. Different from~\cite{deng2020mimamo}, we only use one convolutional-layer block considering training effort. 
% 要對比一下兩者的model parameter區別
% regression module
The output of the two-stream network connects to the regression module, which combines the information of the whole video to achieve frame-level predictions of valence and arousal values. 
% LDS module
The Label distribution smoothing (LDS) module is modified based on DIR~\cite{DIR}.
As validated in~\cite{DIR}, the empirical label distribution does not reflect the real label density distribution in the continuous case. This is because of the dependence between data samples at nearby labels. Fig.~\ref{fig:LDS} illustrates LDS and how it smooths the label density distribution. It convolves a symmetric kernel with the empirical label density to estimate the effective label density distribution that accounts for the continuity of labels
In this work, we re-weight the loss function by multiplying it by the
inverse of the LDS estimated label density for each target.

% \textbf{LDS for Imbalanced Valence Arousal Density Estimation.}

\section{EXPERIMENTS}

\subsection{Dataset}
\subsubsection{Aff-Wild2 dataset}
The Aff-Wild2 is the largest in-the-wild database annotated for valence and arousal. It contains 558 videos with frame-level annotations for valence-arousal estimation, along with facial action unit detection, and expression classification tasks. 
In this work, we only focus on estimating valence and arousal values, which take values in -1 to 1 and -5 represent no annotated values. In the VA set, there are 422 subjects with 1,932,935 images in the training and validation and 139 subjects with 714,986 images in the test. These cropped and aligned images were all provided by ABAW2 ICCV-2021 Competition organizers.
\subsubsection{Aff-Wild dataset}
Aff-Wild dataset is an in-the-wild dataset \cite{kollias2019deep} and contains 298 videos from YouTube, which used the keyword "reaction" to collect. Aff-wild dataset includes valence and arousal annotations ranging continuously in [-1, 1] and has 252 videos and 1,008,652 frames in the training set, and 46 videos and 215,441 frames in the test set.
\subsubsection{AffectNet dataset}
AffectNet \cite{mollahosseini2017affectnet} contains more than 1M images collected by Google, Bing, and Yahoo, and it has valence and arousal, facial action unit, and expression annotations for 450,000 images. Notably, this dataset does not have temporal information, so we only apply it to the model without temporal stream.

%\includegraphics{model.jpg}
%\graphicspath{{./}}
%\DeclareGraphicsExtensions{.jpg}

\subsection{Evaluation metric}
To measure the agreement between the outputs of the model and the ground truth, the Concordance Correlation Coefficient (CCC) metric is used as follow:
\[
CCC=\frac{2\rho\sigma_x\sigma_y}{\sigma^2_x+\sigma^2_y+(\mu_x-\mu_y)^2}
\]
where x and y are the predictions and annotations, $\mu_x$ and $\mu_y$ are the mean values, $\sigma^2_x$ and $\sigma^2_y$ are their variances, and $\rho$ is the correlation coefficient.

The mean value of CCC for valence and arousal estimation will be adopted as the main evaluation criterion of the ABAW2 challenge.
% 如果想增加字数，可以把PCC介绍也放上去，参照AAAI2021论文

% \subsection{Method}
% We try with and without temporal stream two methods. In with temporal stream method, we use Aff-wild and Aff-wild2 datasets. In without temporal stream method, we use Aff-wild, Aff-wild2, and AffectNet datasets. 
\subsection{Experimental settings}
% 各種參數
We implemented our network in PyTorch. The network is trained on NVIDIA GTX 2080Ti with 11GB memory.
% and optimized with the Adam optimizer using default parameters. 
% The inputs are batches of ... clips. We use learning rates with... Weight decay is set to ....
\subsubsection{Data Pre-processing}
% 這段之後要重新整理一下
We merge the training set and validation set and use the cross-validation method to acquire a more accurate estimate of model prediction performance. To apply the Aff-Wild2 dataset, we remove unannotated frames from the beginning and let the remaining ones match its annotation values. 

Notably, the test set may encounter missing frames. We address those missing frames by using two different methods to deal with two situations. If the disregarding frames are at the beginning of the video, we label -5 as the prediction. And if the removed frames are not the case of above, we take the predicted value of the previous frame as its estimation.

\subsection{Experimental Results}
% \subsection{Results and discussion}
% 這部分之後要補充更多實驗，以及對應的discussion
Our valence CCC and arousal CCC are 0.415 and 0.511 on the validation set. We find that the performance on arousal is better than the performance on valence. 
Since arousal describes how active the person is, it should be more related to facial motion than valence.

\subsubsection{Performance Comparison}
% Comparison table，不同方法的比較
In this subsection, we provide the comparison of different methods on the validation set of Affwild2 (see Table~\ref{tab:comparison}). Our spatial stream model achieves 0.591 and 0.617 on valence and arousal respectively.

\begin{table}[!t]
\setlength{\belowcaptionskip}{-5mm}
\centering
\caption{Comparison with other methods on the validation set of Aff-Wild2}
  \begin{tabular}{c|c|c|c}
  \hline
  \multirow{2}{*}{Method} & \multicolumn{3}{c}{CCC} \\\cline{2-4}
  \multicolumn{1}{l|}{}&Valence&Arousal&Mean\\
  \hline
  \hline
  VGG-Face~\cite{kollias2021analysing} & 0.230& 0.210&0.220\\
  \hline
  Affective Process~\cite{CVPR21_sanchez2021affective}& 0.438 & 0.498&0.468\\
  \hline
  \textbf{Ours (spatial)} & \textbf{0.591} & \textbf{0.617}&\textbf{0.604}\\
  \hline
%   $L_{CE} + L_{L2}$ & 86.99 & 87.31\\
%   \hline
  \hline
  \end{tabular}
  \label{tab:comparison}
\end{table}

\subsubsection{Ablation study}
% 有無LDS的對比
To validate the effectiveness of the LDS module, we conduct an ablation study as shown in Table~\ref{tab:LDS ablation study}. Valence and arousal estimation can both benefit from the LDS module. 
This demonstrates that LDS captures the imbalance that affects valence and arousal regression problems.
\begin{table}[!t]
\setlength{\belowcaptionskip}{-5mm}
\centering
\caption{Ablation study of the LDS module on the validation set of Aff-Wild2}
  \begin{tabular}{c|c|c|c}
  \hline
  \multirow{2}{*}{Method} & \multicolumn{3}{c}{CCC} \\\cline{2-4}
  \multicolumn{1}{l|}{}&Valence&Arousal&Mean\\
  \hline
  \hline
  Ours (spatial) without LDS & 0.603 &0.502 &0.552\\
  \hline
  \textbf{Ours (spatial) with LDS} & \textbf{0.604} & \textbf{0.515}&\textbf{0.560}\\
  \hline
%   $L_{CE} + L_{L2}$ & 86.99 & 87.31\\
%   \hline
  \hline
  \end{tabular}
  \label{tab:LDS ablation study}
\end{table}

% 三個loss的對比

% https://heartbeat.fritz.ai/5-regression-loss-functions-all-machine-learners-should-know-4fb140e9d4b0
\section{Conclusion}
We have conducted valence-arousal estimation in the Aff-Wild2 dataset by introducing a two stream learning network. Moreover, we apply label distribution smoothing (LDS) to tackle data imbalanced problem.
Our proposed method achieves Concordance Correlation Coefficient (CCC) of 0.591 and 0.617 for valence and arousal on the validation set of Aff-wild2 dataset.
In the future, we will improve the label distribution re-weight mechanism to achieve better performance.

\bibliographystyle{IEEEtran}
\bibliography{ref}

\begin{thebibliography}{10}
\providecommand{\url}[1]{#1}
\csname url@samestyle\endcsname
\providecommand{\newblock}{\relax}
\providecommand{\bibinfo}[2]{#2}
\providecommand{\BIBentrySTDinterwordspacing}{\spaceskip=0pt\relax}
\providecommand{\BIBentryALTinterwordstretchfactor}{4}
\providecommand{\BIBentryALTinterwordspacing}{\spaceskip=\fontdimen2\font plus
\BIBentryALTinterwordstretchfactor\fontdimen3\font minus
  \fontdimen4\font\relax}
\providecommand{\BIBforeignlanguage}[2]{{%
\expandafter\ifx\csname l@#1\endcsname\relax
\typeout{** WARNING: IEEEtran.bst: No hyphenation pattern has been}%
\typeout{** loaded for the language `#1'. Using the pattern for}%
\typeout{** the default language instead.}%
\else
\language=\csname l@#1\endcsname
\fi
#2}}
\providecommand{\BIBdecl}{\relax}
\BIBdecl

\bibitem{kollias2020analysing}
D.~Kollias, A.~Schulc, E.~Hajiyev, and S.~Zafeiriou, ``Analysing affective
  behavior in the first abaw 2020 competition,'' in \emph{2020 15th IEEE
  International Conference on Automatic Face and Gesture Recognition (FG
  2020)(FG)}, pp. 794--800.

\bibitem{kollias2019expression}
D.~Kollias and S.~Zafeiriou, ``Expression, affect, action unit recognition:
  Aff-wild2, multi-task learning and arcface,'' \emph{arXiv preprint
  arXiv:1910.04855}, 2019.

\bibitem{kollias2019face}
D.~Kollias, V.~Sharmanska, and S.~Zafeiriou, ``Face behavior a la carte:
  Expressions, affect and action units in a single network,'' \emph{arXiv
  preprint arXiv:1910.11111}, 2019.

\bibitem{zafeiriou2017aff}
S.~Zafeiriou, D.~Kollias, M.~A. Nicolaou, A.~Papaioannou, G.~Zhao, and
  I.~Kotsia, ``Aff-wild: Valence and arousal ‘in-the-wild’challenge,'' in
  \emph{Computer Vision and Pattern Recognition Workshops (CVPRW), 2017 IEEE
  Conference on}.\hskip 1em plus 0.5em minus 0.4em\relax IEEE, 2017, pp.
  1980--1987.

\bibitem{kollias2021affect}
D.~Kollias and S.~Zafeiriou, ``Affect analysis in-the-wild: Valence-arousal,
  expressions, action units and a unified framework,'' \emph{arXiv preprint
  arXiv:2103.15792}, 2021.

\bibitem{kollias2021distribution}
D.~Kollias, V.~Sharmanska, and S.~Zafeiriou, ``Distribution matching for
  heterogeneous multi-task learning: a large-scale face study,'' \emph{arXiv
  preprint arXiv:2105.03790}, 2021.

\bibitem{2106.15318}
D.~Kollias, I.~Kotsia, E.~Hajiyev, and S.~Zafeiriou, ``Analysing affective
  behavior in the second abaw2 competition,'' 2021.

\bibitem{deng2020multitask_FG1st}
D.~Deng, Z.~Chen, and B.~E. Shi, ``Multitask emotion recognition with
  incomplete labels,'' in \emph{2020 15th IEEE International Conference on
  Automatic Face and Gesture Recognition (FG 2020)}.\hskip 1em plus 0.5em minus
  0.4em\relax IEEE, 2020, pp. 592--599.

\bibitem{kossaifi2017afew}
J.~Kossaifi, G.~Tzimiropoulos, S.~Todorovic, and M.~Pantic, ``Afew-va database
  for valence and arousal estimation in-the-wild,'' \emph{Image and Vision
  Computing}, vol.~65, pp. 23--36, 2017.

\bibitem{DIR}
Y.~Yang, K.~Zha, Y.-C. Chen, H.~Wang, and D.~Katabi, ``Delving into deep
  imbalanced regression,'' \emph{arXiv preprint arXiv:2102.09554}, 2021.

\bibitem{deng2020mimamo}
D.~Deng, Z.~Chen, Y.~Zhou, and B.~Shi, ``Mimamo net: Integrating micro-and
  macro-motion for video emotion recognition,'' in \emph{Proceedings of the
  AAAI Conference on Artificial Intelligence}, vol.~34, no.~03, 2020, pp.
  2621--2628.

\bibitem{kollias2021analysing}
D.~Kollias, I.~Kotsia, E.~Hajiyev, and S.~Zafeiriou, ``Analysing affective
  behavior in the second abaw2 competition,'' \emph{arXiv preprint
  arXiv:2106.15318}, 2021.

\bibitem{barros2018omg}
P.~Barros, N.~Churamani, E.~Lakomkin, H.~Siqueira, A.~Sutherland, and
  S.~Wermter, ``The omg-emotion behavior dataset,'' in \emph{2018 International
  Joint Conference on Neural Networks (IJCNN)}.\hskip 1em plus 0.5em minus
  0.4em\relax IEEE, 2018, pp. 1--7.

\bibitem{kollias2019deep}
D.~Kollias, P.~Tzirakis, M.~A. Nicolaou, A.~Papaioannou, G.~Zhao, B.~Schuller,
  I.~Kotsia, and S.~Zafeiriou, ``Deep affect prediction in-the-wild: Aff-wild
  database and challenge, deep architectures, and beyond,'' \emph{International
  Journal of Computer Vision}, pp. 1--23, 2019.

\bibitem{chang2017fatauva}
W.-Y. Chang, S.-H. Hsu, and J.-H. Chien, ``Fatauva-net: An integrated deep
  learning framework for facial attribute recognition, action unit detection,
  and valence-arousal estimation,'' in \emph{Proceedings of the IEEE conference
  on computer vision and pattern recognition workshops}, 2017, pp. 17--25.

\bibitem{pan2019deep}
X.~Pan, G.~Ying, G.~Chen, H.~Li, and W.~Li, ``A deep spatial and temporal
  aggregation framework for video-based facial expression recognition,''
  \emph{IEEE Access}, vol.~7, pp. 48\,807--48\,815, 2019.

\bibitem{kim2021contrastive}
D.~H. Kim and B.~C. Song, ``Contrastive adversarial learning for
  person-independent facial emotion recognition,'' 2021.

\bibitem{mollahosseini2017affectnet}
A.~Mollahosseini, B.~Hasani, and M.~H. Mahoor, ``Affectnet: A database for
  facial expression, valence, and arousal computing in the wild,'' \emph{IEEE
  Transactions on Affective Computing}, vol.~10, no.~1, pp. 18--31, 2017.

\bibitem{CVPR21_sanchez2021affective}
E.~Sanchez, M.~K. Tellamekala, M.~Valstar, and G.~Tzimiropoulos, ``Affective
  processes: stochastic modelling of temporal context for emotion and facial
  expression recognition,'' in \emph{Proceedings of the IEEE/CVF Conference on
  Computer Vision and Pattern Recognition}, 2021, pp. 9074--9084.

\end{thebibliography}


%\end{IEEEbiography}


%\newpage



%\vfill


\end{document}